\documentclass{elsarticle}
\usepackage{lineno,hyperref}
\modulolinenumbers[5]
\usepackage{tablefootnote}
\usepackage{multirow}
\usepackage{color}

\usepackage{adjustbox}
\usepackage{enumitem}
\usepackage[T1]{fontenc} %

\begin{document}

\begin{frontmatter}

\title{Wearable-based Mediation State Detection in Individuals with Parkinson's Disease}
\author{Murtadha D. Hssayeni\fnref{Murtadha}}
\fntext[Murtadha]{Department of Computer and Electrical Engineering and Computer Science, Florida Atlantic University, Boca Raton, FL 33431 US.}

\author{Michelle~A.~Burack,~M.D.\fnref{Michelle}}
\fntext[Michelle]{Department of Neurology, University of Rochester Medical Center, Rochester, NY, 14642 US.}

\author{Joohi Jimenez-Shahed,~M.D.\fnref{Joohi}}
\fntext[Joohi]{Department of Neurology, Baylor College of Medicine, Houston, TX US.}

\author{Behnaz~Ghoraani,~Ph.D.\fnref{Murtadha,Behnaz}}
\fntext[Behnaz]{Corresponding Author, Email: bghoraani@ieee.org.}

\begin{abstract}
One of the most prevalent complaints of individuals with  mid-stage and advanced Parkinson's disease (PD) is the fluctuating response to their medication (i.e., ON state with maximum benefit from medication and OFF state with no benefit from medication). In order to address these motor fluctuations, the patients go through periodic clinical examination where the treating physician reviews the patients' self-report about duration in different medication states and optimize therapy accordingly. Unfortunately, the patients' self-report can be unreliable and suffer from recall bias. There is a need to a technology-based system that can provide objective measures about the duration in different medication states that can be used by the treating physician to successfully adjust the therapy. In this paper, we developed a medication state detection algorithm to detect medication states using two wearable motion sensors. A series of significant features are extracted from the motion data and used in a classifier that is based on a support vector machine with fuzzy labeling. The developed algorithm is evaluated using a dataset with 19 PD subjects and a total duration of 1,052.24 minutes (17.54 hours). The algorithm resulted in an average classification accuracy of 90.5\%, sensitivity of 94.2\%, and specificity of 85.4\%.
\end{abstract}


\end{frontmatter}


\section{Introduction}
	Parkinson's disease (PD) is one of the most common chronic progressive neurological disorders, affecting over half a million Americans and resulting in over \$20 billion in direct and indirect costs a year, which is predicted to double by 2040.  \cite{NINDS2015}. PD leads to devastating chronic complications including observable motor symptoms such as tremor, bradykinesia/akinesia (reduced speed and quantity of spontaneous movement), and gait/balance impairment leading to falls, as well as non-motor symptoms such as cognitive impairment and sleep disorders. Levodopa is the most common medication used to improve motor impairments in subjects with PD. Unfortunately, prolonged treatment with Levodopa causes troubling motor fluctuations \cite{Dewey2004}. Almost all patients under the age of 40 will develop motor complications after 6 years from the introduction of Levodopa \cite{Davie2008}. These complications result in frequent fluctuations of "\textit{response to treatment intervention}" between "ON" state with maximum benefit from Levodopa and "OFF" state with least benefit from Levodopa, and are a major focus of PD management \cite{jankovic2005motor}. The current protocol to address these motor fluctuations involves adjusting therapy, e.g., medication frequency and dosage or Deep Brain Stimulation (DBS) parameters, according to durations in medication ON and OFF states that are obtained from patient's self-reports. Patient self-reports require extensive patient education. Even then, self-reports can be unreliable and suffer from recall bias \cite{Maetzler2016}, resulting in frequent clinical examinations to adjust therapy. Unfortunately, these frequent clinical examinations result in an over 50\% increase in just annual direct costs and may not be practical in rural areas where neurologists are not widely available. Technology-based assessments of patients' response to therapy, using wearable sensors, holds great promise in providing objective measures during subjects' daily, free-living condition that can be used by the treating physician to adjust therapy \cite{Rovini2017}. In this work, our objective is to develop a sensor-based assessment system that can detect patients' responses to treatment interventions (i.e., duration in medication ON and OFF states) that are obtained from patients' self-reports. The development of a sensor-based assessment system that can passively detect clinically important information on patients' duration in medication ON and OFF states, from natural environment of individuals, will play a significant role in yielding improved PD therapy adjustment strategies to reduce complicated motor fluctuations and associated healthcare costs without the need for less reliable patient self-reports or frequent clinical examinations. 

	The wide availability of wearable inertial sensors, combined with machine learning algorithms, has led to the development of algorithms for detection of medication ON and OFF states in subjects with PD. However, the existing approaches do not fully address the need for clinically actionable information as required for the therapy adjustments.
	First, some approaches obtained accurate results if the detection was performed for some specific activity, such as walking \cite{perez2015monitoring,rodriguez2015validation,Sama2012} or non-walking \cite{keijsers2006ambulatory}. This is problematic as those methods can identify medication states only during those specific activities and are unable to provide a continuous monitoring of the subjects as needed for detection of clinically important information about duration in different medication states. Second, some approaches trade accuracy for continuous monitoring \cite{hoff2004accuracy,khan2014wearable,salarian2006ambulatory,hammerla2016deep,fisher2016body}, or have to use five to seven sensors at different parts of the body in order to provide acceptable accuracy. As a result, the condition under which the subjects need to use the device is very impractical with too many sensors to wear. Third, these systems are based on a "one-size-fits-all" approach and do not consider inter-subject variability. For example, one subject's ON state could be another subject's OFF state depending on the disease stage and each individual's variability with respect to somatotopy (body parts are impaired), phenomenology (different motor impairments are present), and severity (e.g. one subject's best movement speed may correspond to a less advanced subject's worst movement speed). However, the underlying algorithms of these systems are modeled using data from a group of well-characterized subjects and are used on new subjects \cite{Kubota2016}. Hence, they have limited generalizability and still require additional clinical examinations be performed to customize their report for a new subject. 
	
	In this work, we developed and validated through experiments an algorithm to automatically detect medication ON and OFF states. Our approach is novel in that it is customized to each subject rather than a "one-size-fits-all" approach, and can continuously detect and report medication ON and OFF states as subjects perform different daily routine activities. To attain our objectives, we tested the hypothesis that the medication ON and OFF patterns in two wearable sensor data can be modeled and identified for every individual, when compared to the gold-standard medication ON and OFF states reported by a neurologist. The population of interest in this study is PD patients with motor fluctuation who go through follow-up visits as frequently as every 6 months. There will be multiple adjustments in the medications, which are heavily based on the subjects' self-report. Hence, patients would benefit most from an automated and user-friendly system that is trained in their first visit, and then readily used to detect their response to medication (durations in ON/OFF medication states) on a continuous basis and provide these objective measures that can be used by the treating physician to adjust therapy.

	
	\section{Methods} \label{methods_all}
   We developed a new approach to detect medication ON/OFF states of PD subjects using data from two triaxial gyroscope sensors as patients performed a variety of daily-life activities. The approach consists of a feature extraction and selection stage, which represents each segmented window as a set of features, and a classifier based on a support vector machine (SVM) with fuzzy labeling. The details of the data collection and processing, as well as feature extraction and training the classifier, are explained in this section. 
    \subsection{Dataset}\label{Materials}
\begin{figure}
		\centering
		\includegraphics[width=0.8\columnwidth]{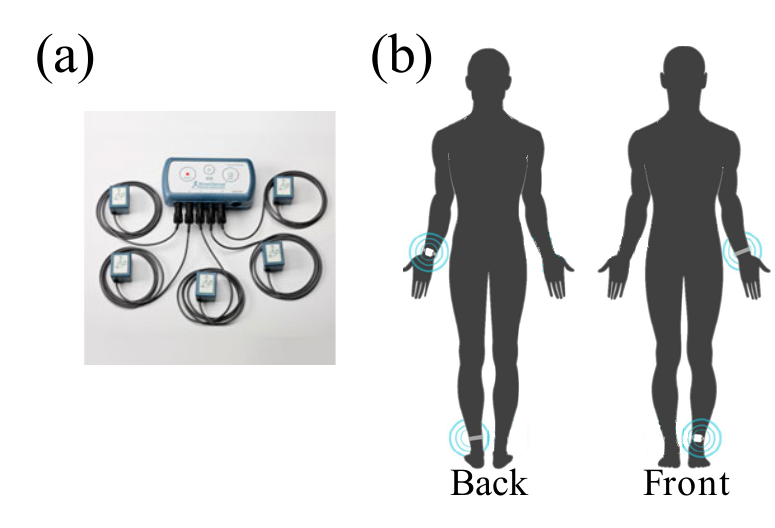}
		\caption{(a) KinetiSense motion sensor unit. (b) The locations of the wearable sensors.} 
		\label{sensors_locations}
	\end{figure}

	A total of 19 PD subjects were selected in this study (see Table \ref{datasets_table} for subject demographics). The recording system (Figure \ref{sensors_locations}(a)) was based on a KinetiSense motion sensor unit (Great Lakes NeuroTechnologies Inc., Cleveland, OH) consisting of a triaxial accelerometer and triaxial gyroscope with 128 Hz sampling rate. Two of these units were mounted on the back and front of the most affected wrist and ankle, respectively, as shown in Figure \ref{sensors_locations}(b). The study was approved by the institutional review board, and all patients provided written informed consent. The dataset was collected under the protocol in Refs. \cite{Mera2013,pulliam2017}. All the participants were asked to stop their medication the night before the scheduled examination and started the experiment in their medication OFF state. First, they were asked to perform the following seven daily living activities: resting, walking, drinking, dressing, hair brushing, unpacking groceries, and cutting food. Next, the subjects took their normal anti-parkinsonian medications and performed the same daily activities again. Later, when the subjects were in their ON state, confirmed by a neurologist's direct observation, they were asked to cycle through all the stations one more time. Concurrently, the clinical examinations were performed by a neurologist to measure and record patients' Unified Parkinson's Disease Rating Scale (UPDRS) \cite{jankovic2005motor} and tremor scores during OFF and ON states. The UPDRS is composed of four subscales: Part 1 and 2 considers non-motor symptoms and motor experiences of daily living, respectively. Part 3 measures severity of motor complications and is performed by clinical examinations, and part 4 measures motor fluctuation and dyskinesia (abnormal involuntary movements) complications. Tremor score for the more affected hand/foot was reported with the following ratings: 0 = absent. 1 = possible rest tremor. 2 = slight and infrequently present. 3 = mild in amplitude and persistent, or moderate in amplitude, but only intermittently present. 4 = moderate in amplitude and present most of the time. 5 = marked in amplitude and present most of the time.

	\begin{table}[ht]
		\centering
		\caption{The Subject Demographics. AC-UPDRS stands for average change between the ON- and OFF-state UPDRS, and LEDD for Levodopa Equivalent Daily Dose. Values are presented as n, mean $\pm$ STD and/or [range].}
		\label{datasets_table}
		\def\arraystretch{1} 
		{\resizebox{\linewidth}{!}{\begin{tabular}{|l|l|l|l|} \hline
				\textbf{Number of Subjects} & 19 & \textbf{OFF UPDRS} &  31 $\pm$ 13 [12-60]   \\ \hline
				\textbf{Age [yr-yr]}            & [42-77]  & \textbf{ON UPDRS} & 14 $\pm$ 8 [4-35]            \\ \hline
				\textbf{Sex (M, F)}    & 14, 5  & \textbf{AC-UPDRS}    & 14.4$\pm$7.9                \\ \hline
				\textbf{Disease Duration (y)}    & 9.2  $\pm$ 3.8  &  \textbf{OFF Tremor (Hand, Foot)}    & 0.68 $\pm$ 1.11, 0.53 $\pm$ 0.84  \\ \hline
				\textbf{LEDD (mg)} & 1282.5$\pm$459.8 & \textbf{ON Tremor (Hand, Foot)}& 0.05 $\pm$ 0.23, 0.00 $\pm$ 0.00\\ \hline
			\end{tabular}}}
	\end{table}
		
		\subsection{Data Preprocessing} \label{preprocessing}
		The KinetiSense motion sensors used in the collection provided two motion data types: gyroscope and accelerometer. We used the gyroscope data as is less affected by vibration and mechanical noise than an accelerometer and is expected to better reflect limb rotations that occur during tremor and other PD motor symptoms due to absence of gravity \cite{Mera2013, fabbrini2007levodopa, Tsipouras2012}. The gyroscope signals were preprocessed by applying a bandpass FIR filter with a pass frequency between (0.5-15Hz) on the three axes of the recorded signal from each sensor to eliminate low and high-frequency noises. Next, the filtered data was segmented into 5-second-long signals with 4-second overlaps between each segments as shown to be a suitable signal duration to detect bradykinesia and tremor \cite{patel2009monitoring}. 
		
		\begin{table} [h]
		\centering
		\caption{The average duration of the data and number of segments that were used for training and testing the algorithm. Values are presented as mean $\pm$ SD. }
		\label{Selecting-data}
		\def\arraystretch{1} 
		\setlength\tabcolsep{4pt} 
		\resizebox{\linewidth}{!}{\begin{tabular}{|c|c|c|c|c|}
				\hline
	\multicolumn{1}{|c|}{}   &    \multicolumn{2}{c|}{\textbf{Training}	} & \multicolumn{2}{c|}{\textbf{Testing}  }          \\ 
    \multicolumn{1}{|c|}{}  & {OFF}	 & {ON}   &{OFF}	 & {ON}   \\  
    \hline
				\textbf{Minutes} 
                & 4.08 $\pm$ 2.61  & 3.80 $\pm$ 2.68   & 37.00 $\pm$ 47.79     & 15.77 $\pm$ 13.30  \\ \hline
				\textbf{Segments} 
                & 223 $\pm$ 149 & 209 $\pm$ 156  & 2,203 $\pm$ 2,880 & 902 $\pm$ 824  \\ \hline
                \textbf{Activities} &
                {\begin{tabular}[l|]{@{}c@{}} part of walking, drinking, \\ resting, and dressing  \end{tabular} } 
                  & {\begin{tabular}[l|]{@{}c@{}}remaining of walking, drinking, \\ resting, and dressing  \\ + \\hair brushing, unpacking groceries, \\ and cutting food \end{tabular} }   & {\begin{tabular}[l|]{@{}c@{}} part of walking, drinking, \\ resting, and dressing  \end{tabular} }     & {\begin{tabular}[l|]{@{}c@{}}remaining of walking, drinking, \\ resting, and dressing  \\ + \\hair brushing, unpacking groceries, \\ and cutting food \end{tabular} }  \\ \hline
    
			\end{tabular}}
		\end{table}

\subsection{Selecting Training and Testing Data}\label{selectingTrainTest}
In a real-life scenario, the developed algorithm will be trained during a patient's first visit. Then, it will be used to detect the response to medication (ON/OFF medication states) on a continuous basis and report objective information about the duration in ON and OFF states to the treating neurologist for remote medication adjustments. We selected the training data such that it is feasible to collect during a routine clinical assessment, and does not cause additional burden on the subjects or their treating physicians. Hence, the training data was selected to be short in duration and also from the activities that can be performed in an office setting (i.e., ambulation, drinking, arm resting, and dressing) and are normally performed as part of the routine assessments in Parkinson's disease (CAPSIT-PD) and Unified Dyskinesia Rating Scale (UDysRS) \cite{goetz2008unified}. The average duration of the ON and OFF training and testing data and the number of analyzed segments are reported in Table \ref{Selecting-data}. Note that there is no overlap between the training and testing data.

		\subsection{Feature Extraction} \label{section:Features}
		 The total number of extracted features for every window, before any feature selection, was 69. Twenty-two of these features were extracted from each X, Y, and Z signals (resulting in a total of 66 for all the three axes) and three features were jointly extracted from the three axes' signals. These extracted features were concatenated into one feature vector to represent the data. An alternative approach would be extracting the features from the magnitude signal; however, this method is unable to retain the directional information of the subject's movement. Table \ref{Table:features} lists the extracted features from a segmented window, $\{W_i\}_{i=1:N_{Tr}}$, where $N_{Tr}$ is the number of windows in a subject's training data. The signal power of A-B Hz band (Features 1-3) was calculated as the summation of the signal powers of the A-B Hz band in the Fourier Domain. The spectral entropy (Feature 5) of the spectrum $\{W_i(f)\}_{i=1:N_{Tr}}$ represented the complexity in each window. The dominant frequency feature (Feature 7) was the dominant harmony of the power spectral density as PD subjects in the OFF state present lower dominant frequency with lower power than in their ON state \cite{weiss2011toward}. 
        Average jerk (Feature 10) was the mean of the second derivative of angular velocity to represent the rate of change. Temporal Shannon entropy (Feature 20) represented the complexity and randomness in the signal as defined below: 
		\begin{equation}
		\{H_{W_i}\}_{i=1:N_{Tr}}=-\sum_{b\in B_i}(p({b})*log_2(p({b})))
		\end{equation}
		where $B_i$ is a histogram bin set in the range (-400,400) with 200 steps for $W_i$, which are found to be effective in this work, and $p({b})$ is the bin $b$ probability in $B_i$. The range of the bin set comes from the 75th and 25th percentile of the angular velocity reading. Gini Index (Feature 21) with a value between 0 and 1 was extracted to measure the moving complexity and was calculated as follows:
		\begin{equation}
		\{G_{W_i}\}_{i=1:N_{Tr}}=1-\sum_{b\in B_i} p({b})^2
		\end{equation}
        
        Sample entropy (Feature 22) was extracted to measure dissimilarity in a given window ($W_i$). It was calculated by first segmenting $W_i$ into vectors of length $m$ and $m+1$ defined as $X_m(\tau)=\{W_i(\tau), W_i(\tau+1), W_i(\tau+2),. . ., W_i(\tau+m-1)\}$ where $\tau$ is the lag between the segments. Second, the Chebyshev distance was calculated between all the vectors of length $m+1$ ($d1$) and of length $m$ ($d2$). If any of these distances was less than a tolerance $r$, then their corresponding counter $M1$ and $M2$ was increased by one, respectively, as shown below:
		\begin{eqnarray}
		&M1=\sum_{j=1}^{N_S-m-1}  \sum_{k=\{j+1|d1\}}^{N_S-m-1} 1 \label{eq:M1}\\
		&M2=\sum_{j=1}^{N_S-m-1}  \sum_{k=\{j+1|d2\}}^{N_S-m-1} 1 \label{eq:M2}
		\end{eqnarray}
where $d1={d[X_{m+1}(j), X_{m+1}(k)]}<r$, $d2={d[X_{m}(j), X_{m}(k)]}<r$, and $d[.,.]$ is the Chebyshev distance. In this work, $m$ and $r$ were set to two and 20\% of the standard deviation of the gyroscope signal, respectively. Finally, sample entropy was computed as follows:
        \begin{equation}
        \{SE_{W_i}\}_{i=1:N_W}=-\log \frac{M1}
		{M2}
        \end{equation}

         Cross-correlation feature ($\rho_{W_i^{A1} W_i^{A2}}$) (Features 23-25) was computed to quantify symmetry between the $A1$ and $A2$ axes of the recorded gyroscope signal. Since PD subjects tend to bend their limbs in the ON state, these features were designed to use symmetry between different axes to differentiate between the ON and OFF states.
         
        \color{black}

        \begin{table}
        \caption{List of extracted features}
		\label{Table:features}
		\def\arraystretch{1.3}
        \begin{tabular}{|l|l|}
        \hline
        \textbf{Feature \#} & \textbf{Feature Description} \\ \hline
        \textbf{1}          & Signal power of 1-4 Hz band \cite{Sama2012,hssayeni2016automatic} \\ \hline
        \textbf{2}          & Signal power of 4-6 Hz band \cite{hssayeni2016automatic,hughes1992accuracy} \\ \hline
        \textbf{3}          & Signal power of 0.5-15 Hz band \cite{hssayeni2016automatic} \\ \hline
        \textbf{4}          & Percentage of the powers for frequencies > 4Hz \cite{keijsers2006ambulatory} \\ \hline
        \textbf{5}          & Spectral entropy ($SH$) \cite{Tsipouras2012} \\ \hline
        \textbf{6}          & Peak in the power spectral density \cite{weiss2011toward, Mera2013}\\ \hline
        \textbf{7}          & Dominant frequency \cite{weiss2011toward} \\ \hline
        \textbf{8}          & Second peak in the power spectral density \\ \hline
        \textbf{9}          & Secondary frequency associated with the second peak \\ \hline
        \textbf{10}         & Average jerk \cite{hssayeni2016automatic} \\ \hline
        \textbf{11}         & Standard deviation ($\sigma$) \cite{khan2014wearable,Tsipouras2012,hssayeni2016automatic} \\ \hline
        \textbf{12}         & Peak-to-peak in the recorded signal \\ \hline
        \textbf{13}         & Mean value for each window ($\mu_{W_i}$) \cite{altun2010comparative} \\ \hline
        \textbf{14}         & Number of autocorrelation peaks \\ \hline
        \textbf{15}         & Sum of autocorrelation peaks \\ \hline
        \textbf{16}         & Lag of the first autocorrelation peak \cite{cole2010dynamic} \\ \hline
        \textbf{17}         & First autocorrelation peak (excluding the peak at the origin) \\ \hline
        \textbf{18}         & Skewness ($\gamma_1$) \cite{altun2010comparative} \\ \hline
        \textbf{19}         & Kurtosis ($k$)\cite{altun2010comparative} \\ \hline
        \textbf{20}         & Temporal Shannon entropy \cite{Tsipouras2012,hssayeni2016automatic} \\ \hline
        \textbf{21}         & Gini Index \cite{hssayeni2016automatic} \\ \hline
        \textbf{22}         & Sample entropy ($SE$) \cite{Chelaru2010} \\ \hline
        \textbf{23}         & Cross-correlation ($\rho_{W_i^X W_i^Y}$) between X and Y axes\\ \hline
        \textbf{24}         & Cross-correlation ($\rho_{W_i^X W_i^Z}$) between X and Z axes \\ \hline
        \textbf{25}         & Cross-correlation ($\rho_{W_i^Y W_i^Z}$) between Y and Z axes \\ \hline
        \end{tabular}
        \end{table}

		\subsubsection{Selection of Significant Features:} \label{section:featureSelection}
		For every training and testing set of data, we applied a statistical analysis approach to the features from the \underline{training data} to select the features that provide a high discrimination between the ON and OFF medication states. First, the normality of distribution of every feature was assessed using the Anderson-Darling test. Next, the statistical significance of the normally distributed features was tested using unpaired t-test \cite{student1908probable}, and for the non-normally distributed features the Wilcoxon rank sum was used. Features with 5\% significance level (i.e., \textit{p}-value < 0.05) between the ON verses OFF medication states were selected.
		The selected features were used in the training and testing of the classifier as explained in the next section.

		\subsection{SVM Classification With Fuzzy Labeling} \label{Section:supervised_SVM}
		Our prior analysis \cite{hssayeni2016automatic} showed that the support vector machine (SVM) was suitable for modeling ON verses OFF medication states and outperformed the well-known clustering classification methods. This observation was in agreement with the study performed in Ref. \cite{khan2014wearable}. Hence, in this work, the well-known SVM approach was employed. The algorithm consists of a training and a testing stage, where the former adjusts the SVM parameters based on the training data and the latter tests the trained SVM model on the testing data.  
		
		\subsubsection{Training:} \label{TrainingModel}
		We used the recursive feature elimination (REF) method to jointly select features and train SVM hyperparameters \cite{Weston2001}. REF selects features via a greedy backward selection: it trains the SVM hyperparameters, and obtains a feature weight vector based on minimizing generalization bounds of the leave-one-out cross validation, using a gradient descent approach. It then removes the feature with the smallest returned absolute-value weight. This process is repeated until a maximum classification accuracy is achieved. We used the Feature Selection Library MATLAB Toolbox \cite{Roffo2016} to perform the above feature selection process and train the SVM hyperparameters (i.e., linear or RBF kernel, cost parameter ($c \in{2^{\{-2,...,2\}}}$), and gamma parameter ($\gamma \in 2^{\{-4,...,4\}}$)). 
	
\begin{figure}
			\centering
			\includegraphics[scale=0.6]{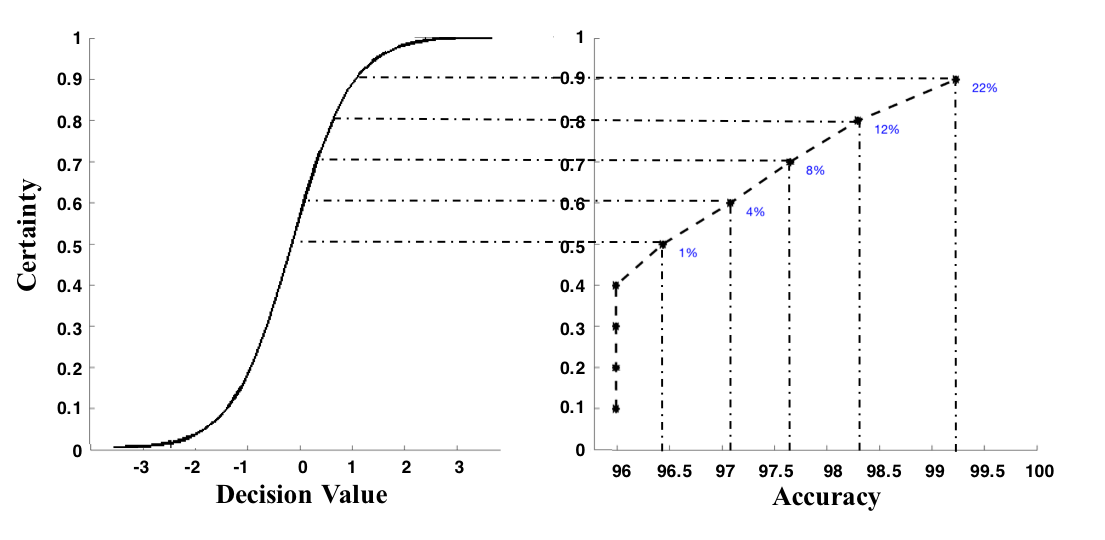}
			\caption{The plot on the left side is a sample of Platt sigmoid and the plot on the right side is the training accuracy corresponding to different threshold values on the classification certainty. The percentages represent the inconclusive data for nine different threshold values from 50\% to 90\%.}
			\label{Figure:Certainty_Threshold_vs_Accuracy}
		\end{figure}

\noindent		\underline{Classification Certainty ($P$):} It is known that SVM does not provide any posterior probability that can represent the model classification certainty. Platt \textit{et al.} \cite{platt1999probabilistic} proposed a parametric form of sigmoid to provide a measure of posterior probability for SVM that measures the confidence of the trained SVM model in detecting the medication state of each window. 
This approach uses a parametric model according to Eq. \ref{eq:platt1} to fit posterior $P(\textbf{F}=1|D)$ where \textbf{F} indicates feature vectors, $D$ represents the SVM decision values, and $1$ refers to the class labels being medication "ON state". The parameters $A$ and $B$ are fit using maximum likelihood estimation from a union of $n$-fold sets of $D$ values from the training data. Each of these sets were obtained after training a SVM model on ($n$-1) fold of the training data and validating them on the remaining fold to find $D$ values for each feature vector in that fold. The union of all $D$ values as a result of the cross-validation on the training data were used as the training set of the sigmoid. Two scalar parameters ($A$ and $B$) were estimated by minimizing a negative log likelihood based on the ground truth labels and the training decision values, $D_{1 \times N_{Tr}}$ \cite{platt1999probabilistic}. We determined these values, $D_{1 \times N_{Tr}}$, from the \underline{training data} based on an $4$-fold cross-validation of the four activities in the training data. The estimated parameters ($A$ and $B$) were then used in Eqs. \ref{eq:platt1} and \ref{eq:platt2} to provide a measure of the classification certainty $ 0\le P_n \le 1$ of a classification decision value $M_n$.
  \begin{eqnarray}
		&\{P_n\}_{\{n=1:N_{Te}|M_n>0\}}=\frac{1}{1+e^{A\times M_n+B}} \label{eq:platt1}\\
		&\{P_n\}_{\{n=1:N_{Te}|M_n<0\}}=\frac{e^{A\times M_n+B}}{1+e^{A\times M_n+B}} \label{eq:platt2}
		\end{eqnarray}	
where a higher value of $P_n$ indicates a higher certainty in the classification decision value. Hence, the decision values with the classification certainty of less than a threshold $Th_{Tr}$ can be reported inconclusive and censored from the medication detection report. Figure \ref{Figure:Certainty_Threshold_vs_Accuracy} displays one example of the change in the percentage of inconclusive data with respect to thresholds 50\% to 90\% on the classification certainty. We selected threshold $Th_{Tr}$ for every training data such that there was at most 1\% rejection rate in the decision report. The threshold $Th_{Tr}$ along with the estimated parameters $A$ and $B$ were then used in the testing phase. 
		\color{black}
		\subsubsection{Testing:}
		The selected features (from the above process) were extracted from the testing data, and passed through the trained SVM model to obtain the decision values ($D_{1 \times N_{Te}}$). Given that the medication state does not normally change so quickly, rapid fluctuating outliers in the predicted decision values were adjusted accordingly. A fuzzy labeling approach was used to adjust the classification label of every second data based on a group of its contiguous feature vectors. This fuzzy labeling process was performed using two averaging filters. The first averaging filter was applied to the decision values ($D_{1 \times N_{Te}}$). The filter had a width of five steps and four-step overlap to reduce the effect of outliers in the contiguous decision values that share part of the signal between each other. The output of this filter was then furthered processed using a sliding 40-second averaging filter. After filtering, the averaged decision values ($M_{1 \times N_{Te}}$) were used to predict the medication state of every window. The medication state was predicted as medication OFF if $M_n$ > zero, and otherwise it was predicted as medication ON. The trained $A$ and $B$ parameters from the training phase were employed to estimate the classification certainty for every second $n$. Eq. \ref{eq:platt1} was used for medication OFF detection cases and Eq. \ref{eq:platt2} was used for cases with medication ON detection. The trained threshold $Th_{Tr}$ was used to select and report the decision values with a high classification certainty value and the remaining data with a certainty value of less than $Th_{Tr}$ was reported as inconclusive.

		
		\section{Results} \label{results}
\begin{figure} [h]
			\centering
			\includegraphics[width=\linewidth]{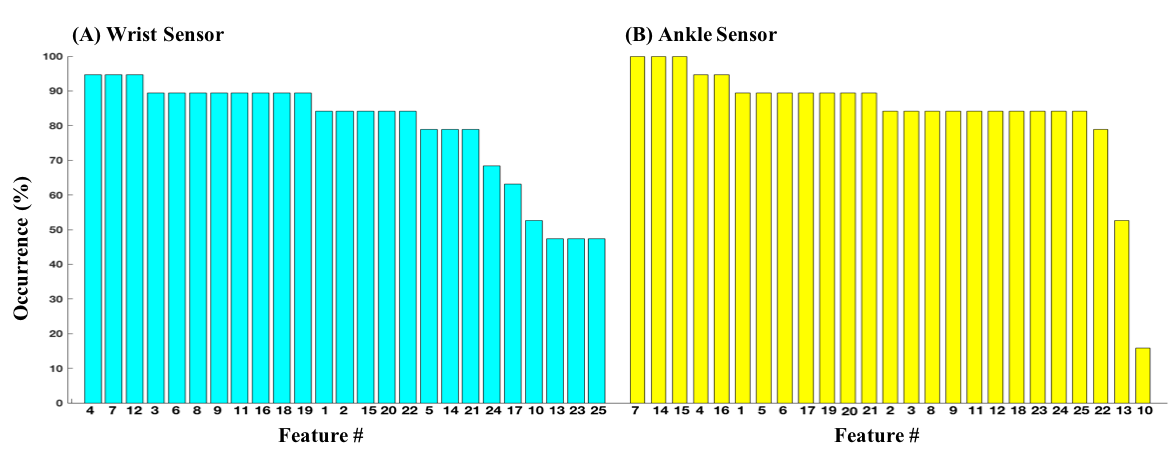}
			\caption{Occurrence rate of the selected features. The features' numbers are as defined in Table \ref{Table:features}.}
			\label{Figure:Feature_Selection}
		\end{figure}

		The developed medication state detection algorithm was trained and evaluated using the dataset in Section \ref{Materials}. The feature selection and SVM training process resulted in different SVM hyper-parameters and features for different subjects. Linear kernel was selected for five individuals and RBF kernel for the rest of them. Figure \ref{Figure:Feature_Selection} shows the occurrence rate of the selected features among all the subjects in our study. A occurrence rate of 100\% means that that specific feature provides a high medication ON/OFF discrimination and was selected for all the subjects.
The testing accuracy, sensitivity, and specificity results of all the subjects are reported in Table \ref{Table:patient_specificResults}. On average, there was 2.7\% of inconclusive data with an average classification accuracy of 90.5\%, sensitivity of 94.2\%, and specificity of 85.4\%. 
		\begin{table}[tbp]
			\centering
			\caption{Testing Results of ON/OFF Medication State Detection.}
			\label{Table:patient_specificResults}
			\def\arraystretch{1}
			\resizebox{\textwidth}{!}{\begin{tabular}{|l|c|c|c|c|c|c|c|c|c|c|}
				\hline
		   &		\multicolumn{10}{|c|}{\textbf{Subjects}}  \\ \cline{1-11}
		\multirow{ 1}{*}{\textbf{Performance}} 	&	\textbf{1}  & \textbf{2} & \textbf{3} & \textbf{4} & \textbf{5} & \textbf{6} & \textbf{7} & \textbf{8} &\textbf{9} & \textbf{10} \\ \hline
		\textbf{Accuracy (\%)}& 99.8 & 99.5 & 96.5 & 95.7 &  95.4 & 94.1&  93.5&  93.4 & 92.8 & 92.5 \\ \hline
   \textbf{Sensitivity (\%)}& 100 & 100 & 99.5& 99.8 & 99.2 & 99.9 & 98.4  &97.3 &95.2 & 96.0     \\ \hline
   \textbf{Specificity (\%)}& 98.7 & 98.1 & 88.8 & 87.6 & 86.3 &  85.4& 85.5 &84.2 &85.8 &  84.9  \\  \hline
   \multicolumn{11}{c}{\textbf{}} \\ \hline
     &		\multicolumn{10}{|c|}{\textbf{Subjects}}  \\ \cline{1-11}
		\multirow{ 1}{*}{\textbf{Performance }} 	&	\textbf{11}  & \textbf{12} & \textbf{13} & \textbf{14} & \textbf{15} & \textbf{16} & \textbf{17} & \textbf{18} &\textbf{19} & \textbf{Ave.} \\ \hline
		\textbf{Accuracy (\%)}&90.7  & 89.3 & 87.4 & 86.5 & 85.6 & 85.2 & 84.7 &79.8 & 77.3 & \textbf{90.5}  \\ \hline
   \textbf{Sensitivity (\%)}& 92.1 & 91.0 & 88.2 & 92.3 & 86.4 & 88.9 &88.3  & 89.4& 88.5 & \textbf{94.2}     \\ \hline
   \textbf{Specificity (\%)}&86.3  & 85.9 & 86.3 & 79.1 & 84.9 & 82.7 & 81.9 & 76.5 & 74.2 & \textbf{85.4}   \\  \hline
			\end{tabular}}
		\end{table}	
		
\begin{figure}[tbp]
			\centering
			\includegraphics[width=\linewidth]{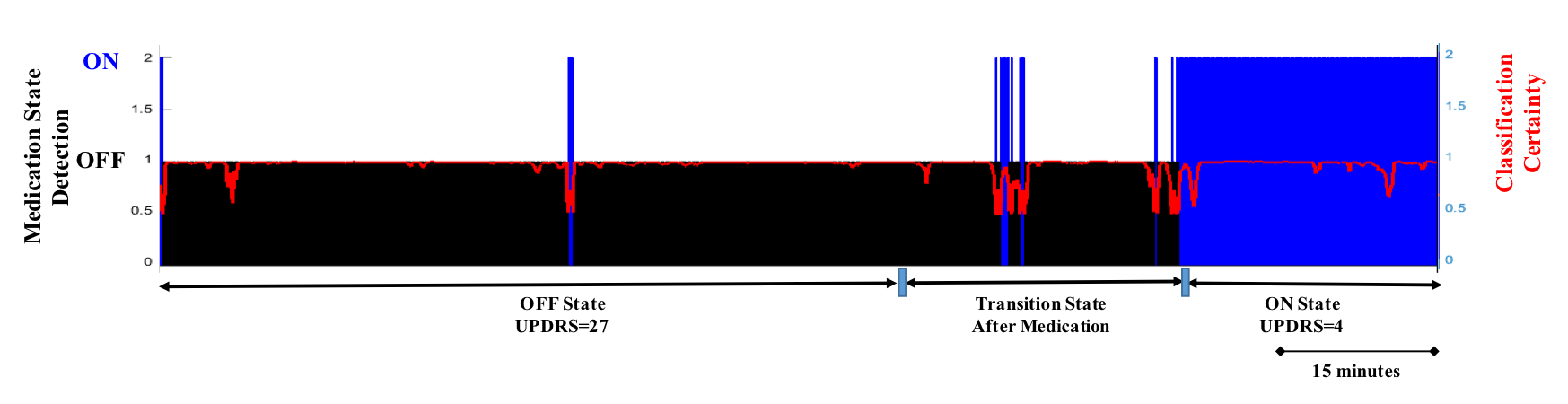}
			\caption{The classification results (1 medication OFF, 2 medication ON) with the certainty (continuous line between 0 and 1) for subject 14.}
\label{Figure:patient14_resultVstime_using_leg}
		\end{figure}

		A sample of the medication ON/OFF report that was predicted by the algorithm is shown in Figure  \ref{Figure:patient14_resultVstime_using_leg}. The tall boxes indicate where a medication ON state was predicated and the shorter boxes show a medication OFF prediction. The red tracing provides the certainty measure of the SVM algorithm with respect to the predicted medication state. A classification certainty of 1 indicates that the SVM classifier makes a prediction with 100\% certainty. As indicated by the red tracing, the algorithm detected the medication states with a high certainty for most of the times with the highest number of false detection occurring during the transition from the OFF to ON state. 

\subsection{Classification Performance and Activity Type}
\begin{figure}[h]
			\centering
			\includegraphics[width=\linewidth]{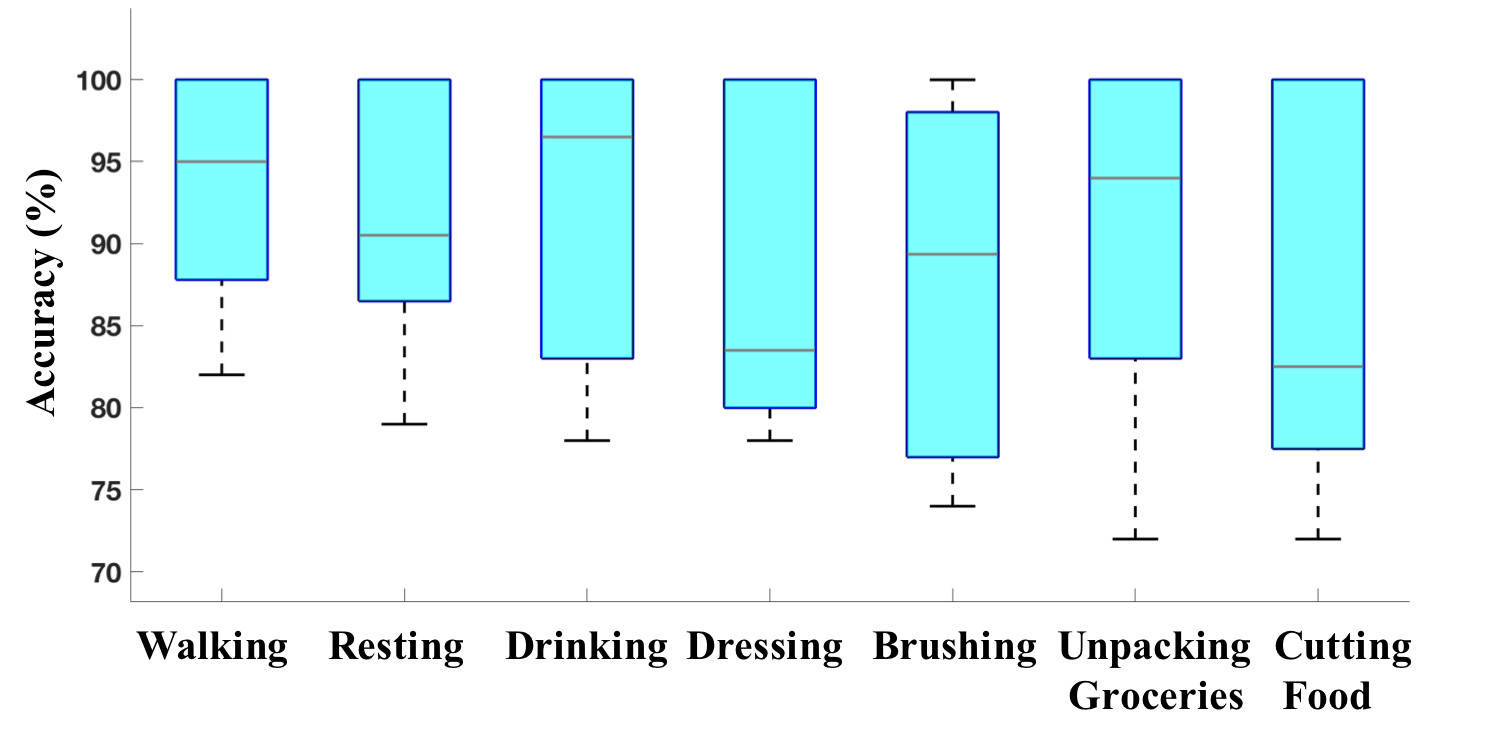}
			\caption{Testing accuracy for different activities. The central mark indicates the median, and the bottom and top edges of the box indicate the 25th and 75th percentiles, respectively. The whiskers extend to the minimum and maximum data points.}
			\label{fig_new}
		\end{figure}

A mediation state monitoring algorithm that can provide accurate and objective information about the duration in ON and OFF states has to continuously detect medication ON/OFF states as the subjects perform a variety of daily routine activities. Here the developed algorithm was trained using data from four activities (i.e., walking, resting, drinking, and dressing) and tested on a total of seven activities consisting of four training activities (with no overlap between the training and testing data) and three additional activities (i.e., brushing, unpacking groceries, and cutting food). To evaluate the robustness of the developed algorithm to the activities that were not included in the classifier training phase, we obtained the average testing accuracy of the algorithm for separate activities (see Figure \ref{fig_new}). The average accuracy for the group of activities in the training phase was 91.3\% and for the group of the new activities, it was 88.4\%. We performed a paired t-test between the average accuracy of the two activity groups as well as the accuracy of all the 21 possible pairs of activities (e.g., walking and brushing, resting and brushing, and etc.). These analyses did not identify any statistically significant difference between classification performance of any of the activities indicating that the algorithm was robust to the new activities that were not used in the training phase. 

\subsection{Sensor Locations and Tremor}     
\begin{figure}
		\centering
		\includegraphics[scale=1]{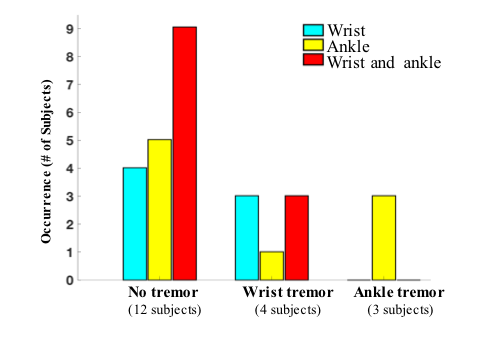}
		\caption{Occurrence of the sensor combinations with the highest testing accuracy for three symptom groups of no tremor, wrist tremor, and ankle tremor.}
		\label{Figure:FistOccurrence_tremor}
	\end{figure}

Fewer number of wearable sensors in a wearable-based mediation state monitoring system can improve its user-acceptability and practicality. Hence, we investigated if there was a relationship between the number and location of the sensors and a disease symptom (here tremor) that can help to minimize the number of sensors while achieving the highest classification accuracy. For this purpose, we repeated the classification training and testing phase as described in Section \ref{Section:supervised_SVM}, using only wrist sensor and only ankle sensor, and calculated the testing accuracy for every subject. Next, the subjects were divided into three symptom groups as diagnosed by the neurologist: without tremor (12 subjects), with dominant tremor on wrist (4 subjects), and with dominant tremor on ankle (3 subjects). We compared the classification accuracy (rounded to the nearest one) of three sensor combinations (both wrist and ankle sensors, only wrist sensor, and only ankle sensor) for every symptom group, and counted the number of subjects with the highest accuracy when using a sensor combination. If one subject had a highest classification accuracy using more than one of the sensor combinations, we counted that subject towards all of those sensor combinations. For example, considering the no-tremor group with 12 subjects, our finding was as follows: Nine subjects had their highest accuracy using both wrist and ankle sensors; however, one of these subjects resulted in an equally-well accuracy when using only wrist sensor, and two of them achieved an equally-well accuracy when using only ankle sensor. Three subjects achieved the best classification accuracy only when wrist sensor was used, and three subjects resulted in their highest classification accuracy only using ankle sensor. These results were counted as 4, 5, and 9 for the best performance using only wrist, only ankle, and both wrist and ankle, respectively. The same process was repeated for the other two symptom groups and the results are reported in Figure \ref{Figure:FistOccurrence_tremor}. One interesting observation from Figure \ref{Figure:FistOccurrence_tremor} is that for all the subjects with ankle tremor, using only ankle sensor was sufficient to achieve the highest classification performance.

\section{Discussions}\label{Discussions}
We developed a technology-based assessment of PD subjects' medication ON and OFF states, using data from two wearable sensors, to provide objective measures, instead of patient diary or self-report, that can be used by the treating physician to adjust therapy \cite{Rovini2017}. We developed a feature extraction and selection algorithm based on SVM classifier with fuzzy labeling. Using concurrently recorded sensor data and gold-standard medication ON and OFF states from 19 PD subjects, we tested our hypothesis that the medication ON and OFF patterns in the data from two wearable sensors can be modeled and identified for every individual. Several interesting observations were made via our analysis: First, our analysis supports our hypothesis by demonstrating that, using the data from two sensors, the developed algorithm was able to predict correct medication states with a high average accuracy of 90.5\%, sensitivity of 94.2\%, and specificity of 85.4\% (Table 4). Moreover, the algorithm performs equally-well for all the seven activities that were considered in this study (Figure 5). Since we trained the classifier using only part of the data from four of the activities and tested for all the data, including three new activities (Table 2), this observation indicates that the algorithm successfully models the medication state patterns, and can potentially be used to detect medication states during subjects' daily routine activities. 

Another interesting observation is that our analysis indicates that the ON/OFF-discriminant features are different between subjects supporting the need for an individualized classifier training approach verses a "one-size-fits-all" approach. As shown in Figure \ref{Figure:Feature_Selection}, there is a difference in the set of features that are selected for every subject and from each sensor. Most of the features were selected for more than 70\% of the subjects, but some of them (six of the wrist features and two of the ankle features) were selected for less than 70\% of the subjects. Dominant "wrist features" with the occurrence of greater than 90\% were the signal power at frequencies > 4Hz \cite{keijsers2006ambulatory}, dominant frequency \cite{weiss2011toward}, and peak-to-peak of the signal. The significant "ankle features" were the number of the autocorrelation peaks and the lag of the first autocorrelation peak. The average jerk and mean features had the least occurrence and the cross-correlation feature was informative only in the case of the ankle sensor.

We also investigated the possibility of using only one sensor for detecting the medication ON/OFF states. For this purpose, we determined the sensor combination that achieved the highest classification accuracy for every subject and provided its distribution for the presence of tremor in the subject's wrist or ankle (Figure 6). Our analysis indicates that different sensor numbers and placements provide different performances. It is interesting that using both wrist and ankle sensors does not always result in the best performance. All the three individuals with the most dominant tremor on their leg had the best classification accuracy using only a single ankle sensor (see Figure 6) and three (out of four) individuals with most dominant tremor on their hand had the best classification accuracy using only a wrist sensor. This observation suggests the possibility of a \textit{strategy in selecting the number and placement of sensors}, which is using only an ankle sensor for individuals with the most dominant tremor on the leg and using an ankle and wrist sensor combination for the other subjects. The reliability of such as sensor placement strategy has to be further investigated in future work with higher number of subjects.

			\subsection{Comparison to Other Studies}            
			A summary of the previous studies for the classification of PD medication states is listed in Table \ref{Table:methods_review}. There are two main types of medication detection approached in the literature. One approach provide medication states only during some specific activities, such as walking or non-walking, \cite{keijsers2006ambulatory,Sama2012,perez2015monitoring, rodriguez2015validation}; however, this approach is unable to provide a continuous monitoring of the subjects as needed for detection of the duration of different medication states. The other approach, similar to our algorithm, makes a medication state detection independent of the activity type \cite{khan2014wearable, salarian2006ambulatory, hammerla2016deep, fisher2016unsupervised,hoff2004accuracy,Um2018}, and hence is able to provide duration in different medication states. Our developed algorithm provided the highest sensitivity (94\%) and specificity (85\%) compared to the state-of-the-art using only two sensors. The second-highest performance was achieved by the work reported in Ref.  \cite{salarian2006ambulatory} with a sensitivity of 90\% and specificity of 76\%, using five sensors, for 13 subjects with an average change between the ON- and OFF-state UPDRS (AC-UPDRS) $\ge$ 16. In our work, the average classification performance for the subjects who had an AC-UPDRS $\ge$ 16 (eight out of 19 subjects) was 98\% sensitivity and 90\% specificity, which is significantly higher that the reported performance in \cite{salarian2006ambulatory}. However, it is worth mentioning that there is a high variability between the different studies in terms of the number of sensors (range of one to seven), data duration (range of one hour to one week), and subject numbers (range of 12 to 19), which makes it difficult to perform a fair comparison between the methods.  
			\begin{table}[tbp]
				\centering
				\caption{Comparison of Medication ON/OFF State Detection Methods}
				\def\arraystretch{1.1} 
				\setlength\tabcolsep{2pt} 
				\label{Table:methods_review}
				\resizebox{\textwidth}{!}{\begin{tabular}{|c|c|c|c|c|c|c|c|}
						\hline
						\textbf{References}                                                           & \textbf{\# Sensors}                                                                                                & \textbf{\# Subjects} & \textbf{\begin{tabular}[c]{@{}c@{}}Data Duration\\ for Each Subject\end{tabular}}            & \textbf{\begin{tabular}[c]{@{}c@{}}Classification\\ Method\end{tabular}}                                  &                                 \textbf{\begin{tabular}[c]{@{}c@{}}Activity-Independent\\ Model (Yes, No)\end{tabular}} & \textbf{\begin{tabular}[c]{@{}c@{}}Individualized-Model\\ (Yes, No, Partial)\end{tabular}}   
						& \textbf{Results}                                                                                                    \\ \hline
						
						Keijsers \textit{et al.} \cite{keijsers2006ambulatory}                                                      & \begin{tabular}[c]{@{}c@{}}One tri-axial\\  accelerometer\end{tabular}                                             & 23                   & 3 hours                                                                                & \begin{tabular}[c]{@{}c@{}}Linear discriminant \\ and ANN\end{tabular}                                    &
						\begin{tabular}[c]{@{}c@{}}No\\ (excludes walking)\end{tabular} &
						\begin{tabular}[c]{@{}c@{}}Partial (Tremor, \\ and non-tremor)\end{tabular}    &   \begin{tabular}[c]{@{}c@{}}Sens.: 97\%\\ Spec.: 97\%\end{tabular}                                                    \\ \hline
						
						Sama \textit{et al.} \cite{Sama2012}                                                          & \begin{tabular}[c]{@{}c@{}}One tri-axial\\ accelerometer\end{tabular}                                              & 20                   & 1 hour                                                                                 & \begin{tabular}[c]{@{}c@{}}SVM and \\ linear discriminant\end{tabular}                                    & No (gait)                                                                                                                             & No                                                                                     & \begin{tabular}[c]{@{}c@{}}Acc.: 94\%, Sens.:\\ 84\%, Spec.: 90\% \end{tabular}                                      \\ \hline
						
						Perez-Lopez \textit{et al.} \cite{perez2015monitoring}                                                 & \begin{tabular}[c]{@{}c@{}}One tri-axial\\ accelerometer\end{tabular}                                              & 7                    & 6 hours                                                                                & Linear discriminant                                 &   \begin{tabular}[c]{@{}c@{}}No (walking and \\ non-walking)\end{tabular}        &
						\begin{tabular}[c]{@{}c@{}}Partial (Threshold \\ on Bradykinesia )\end{tabular} &
						\begin{tabular}[c]{@{}c@{}}Sens.: 99.9\%\\ Spec.: 99.9\%\end{tabular}                                               \\ \hline
						\begin{tabular}[c]{@{}c@{}}Rodriguez-\\ Molinero \textit{et al.} \cite{rodriguez2015validation}\end{tabular} & \begin{tabular}[c]{@{}c@{}}One tri-axial\\ accelerometer\end{tabular}                                              & 35                   & \begin{tabular}[c]{@{}c@{}}1.4 - 5.5 hours\end{tabular}                              & Linear discriminant                                                                                       & No (only walking)                                                                                                                          & Yes                                                                                   & \begin{tabular}[c]{@{}c@{}}Sens.: 96\%\\ Spec.: 94\%\end{tabular}                                                   \\ \hline             
						
						Khan \textit{et al.} \cite{khan2014wearable}                                                          & \begin{tabular}[c]{@{}c@{}}One tri-axial\\ accelerometer\end{tabular}                                              & 12                   & 1 hour                                                                                 & SVM                                                                                                       & Yes                                                                                                                                     & No                                                                                     & Acc.: 72\%                                                                                                          \\ \hline                        
						Salarian \textit{et al.} \cite{salarian2006ambulatory}                                                     & \begin{tabular}[c]{@{}c@{}}Five gyroscopes\\ and accelerometers\end{tabular}                           & \begin{tabular}[c]{@{}c@{}}13\\ AC-UPDRS $\geq$ 16  \end{tabular}                  & \begin{tabular}[c]{@{}c@{}}3-6 hours\end{tabular}                                & Logistic regression                                                                                       & Yes                                                                                                                                     & No                                                                                     & \begin{tabular}[c]{@{}c@{}}Sens.: 90\%\\ Spec.: 76\%\end{tabular}                                               \\ \hline
						Hammerla \textit{et al.} \cite{hammerla2015pd}                                                      & \begin{tabular}[c]{@{}c@{}}Two tri-axial\\ accelerometers\end{tabular}                                              & 32                   & \begin{tabular}[c]{@{}c@{}}4 hours (in lab)\\ 1 week (in home)\end{tabular}                  & \begin{tabular}[c]{@{}c@{}}Restricted \\ Boltzmann Machines\end{tabular}                  & Yes                                                                                                                                    & No                                                                                     & \begin{tabular}[c]{@{}c@{}}Mean f1-score:\\ 68\%\end{tabular}             \\ \hline
						Fisher \textit{et al.} \cite{fisher2016unsupervised}                                                       & \begin{tabular}[c]{@{}c@{}}Two tri-axial\\ accelerometers\end{tabular}                                              & 32                   & \begin{tabular}[c]{@{}c@{}}4 hours (in lab)\\ 1 week (in home)\end{tabular}                  & ANN                                                                                                       & Yes                                                                                                                                     & No                                                                                     & \begin{tabular}[c]{@{}c@{}} Sens.: 55\%\\ Spec.: 83\%\end{tabular} \\ \hline
												Um \textit{et al.} \cite{Um2018}                                                          & \begin{tabular}[c]{@{}c@{}}One \\ accelerometer\end{tabular}                                              & 30                   & 3-4 hours                                                                                     & \begin{tabular}[c]{@{}c@{}}Convolutional \\ Neural Networks\end{tabular}                                            & Yes                                                                                                                                     & No                                                                                    & \begin{tabular}[c]{@{}c@{}}Acc.: 63.1\% for ON, OFF\\ and ON with dyskinesia \end{tabular}                                         \\ \hline
						Hoff \textit{et al.} \cite{hoff2004accuracy}                                                          & \begin{tabular}[c]{@{}c@{}}Seven uni-axial \\ accelerometers\end{tabular}                                              & 15                   & 24 hours                                                                                     & \begin{tabular}[c]{@{}c@{}}Linear \\ discriminant\end{tabular}                                            & Yes                                                                                                                                     & Yes                                                                                    & \begin{tabular}[c]{@{}c@{}}Sens.: 60\%-71\%\\ Spec.: 66\%-76\%\end{tabular}                                         \\ \hline
						\multirow{ 1}{*}{Proposed}                                                                          & 
					\begin{tabular}[c]{@{}c@{}} Two tri-axial \\ gyroscopes\end{tabular} & 19                  & 1 hour & \begin{tabular}[c]{@{}c@{}}SVM with fuzzy\\ classification\end{tabular}                                 &\multirow{ 1}{*}{Yes}                                                                                                                                     & \multirow{ 1}{*}{Yes}                                                                                     & \begin{tabular}[c]{@{}c@{}}Acc.:\textbf{91\%}, Sens.:\\ \textbf{94\%}, Spec.: \textbf{85\%}\end{tabular}       \\
						\hline
					\end{tabular}}
				\end{table}

\subsection{Clinical Implications and Study Limitations}  
The developed algorithm has the potential to establish a wearable-based mediation state monitoring system to provide objective measures, from the PD subjects in free-living condition, that can be used by the treating physician to adjust therapy. We used an individualized approach to train a customized classifier for every subjects. Similar to the other individualized approaches \cite{rodriguez2015validation,hoff2004accuracy,keijsers2006ambulatory,perez2015monitoring}, our work uses labeled training data from the same subject, assuming that some gold-standard medication-state data is available for a subject. Given that the focus of this study is the PD subjects with motor fluctuation who go through frequent follow-up visit, we designed the classifier training phase based on a short duration of training data from the activities (i.e., ambulation, drinking, resting, and dressing) that can be easily performed in any clinical office. Hence, once the algorithm is trained, it can be readily used as a passive system to monitor medication fluctuations without relying on patient or physician engagement. The algorithm was tested using sensor data as the subjects performed several routine activities in a clinical lab setting. In our future work, we will investigate the algorithm's performance as the subjects perform a variety of activities in their homes. We will also investigate the sensitivity of the algorithm to the variations in the sensor placements when the subjects wear their own sensors at home. Our analysis on selecting the number of sensors and their placement was based on only the tremor symptom. The reliability of this selection to optimize the number of required sensors will be studied using more subjects in our future work. 
			
\section{Conclusion}
\label{conclusion}
In this paper, we developed a sensor-based assessment system that can continuously detect patients' medication ON and OFF states as the patients perform different types of daily routine activities. A series of novel signal features and feature selection approaches were proposed and integrated into an SVM classifier with fuzzy labeling. The developed algorithm was assessed using data from 19 PD subjects, and the results supported our hypothesis that it was possible to model and detect the subjects' medication ON and OFF state patterns from the data from two wearable sensors as the subjects perform different activities. The proposed algorithm used data from only two sensors on wrist and ankle, and it was able to detect the response to medication during subjects' daily routine activities with an average 90.5\% accuracy 94.2\% sensitivity, and 85.4\% specificity. The developed algorithm would enable a sensor-based assessment system that can provide objective measures about the duration in medication ON and OFF states that are obtained from patients' self-reports. Such a system could provide accurate and timely therapy adjustments in place of less reliable patient self-reports, and as a result considerably improve both the care delivery and quality of life for the millions of patients afflicted by this debilitating neurodegenerative disorder.


\section*{Acknowledgment}
The dataset was supported by Small Business Innovation Research grant offered by National Institutes of Health (NIH) to Cleveland Medical Devices (1R43NS071882-01A1; T. Mera, PI) and the National Institute on Aging of NIH grant to Great Lakes NeuroTechnologies Inc. (5R44AG044293).

\section*{Conflict of Interest Statement}
None.

\bibliography{mybibfile}

\end{document}